# An Introduction to Person Re-identification with Generative Adversarial Networks

Hamed Alqahtani[1], Manolya Kavakli-Thorne[2], and Charles Z. Liu[3]

[1] Computer Science Dept, King Khalid University
hsqahtani@kku.edu.au
[2] Academic Director, Academy of Interactive Entertainment
[3] VISOR, Macquarie University

**Abstract.** Person re-identification is a basic subject in the field of computer vision. The traditional methods have several limitations in solving the problems of person illumination like occlusion, pose variation and feature variation under complex background. Fortunately, deep learning paradigm opens new ways of the person re-identification research and becomes a hot spot in this field. Generative Adversarial Nets (GANs) in the past few years attracted lots of attention in solving these problems. This paper reviews the GAN based methods for person re-identification focuses on the related papers about different GAN based frameworks and discusses their advantages and disadvantages. Finally, it proposes the direction of future research, especially the prospect of person re-identification methods based on GANs.

## 1 Introduction

Person re-identification is considered as a significant part of multiple cameras pedestrian tracking. It is defined as the process of identifying whether to pedestrian images of disjoint and non-overlapping cameras at different time intervals are the same or not. A typical person re-identification system has three phases, namely, detecting a person, tracking the person and retrieving the person. This work focuses on a review of person retrieval using generative adversarial networks. Several researchers have paid special attention to the problem of person re-identification in the field of computer vision. In spite of much research in this field, it remains a challenging problem for the researchers in the field.

It has been observed that most of the person images are clicked by disjoint and non-overlapping cameras installed in an uncontrolled environment, having a low quality of the images. The low-quality images it difficult to use by the conventional systems for extracting their features to be used for detecting persons' faces accurately. So, the features representing the appearance of the person must be extracted on the basis of clothes colours or some object with the person to identify them in different poses of multiple cameras. Therefore, these appearance features are more suitable for a person re-identification process. However, the appearance features are helpless in case of similar cloth colours. In addition, the appearance of persons also changes in multiple camera views due to lightning change in different background poses. It may also be possible that different persons have similar appearances or resemblance. Multiple camera views also change with the passage of time and movement of the person.

Person re-identification is considered as an important task in the field of computer vision. Several research efforts have been made for effective person re-identification. The significant existing methods can be divided into two categories, namely, traditional methods and deep learning methods [9][13] [29]. Traditional methods involves 1) Extracting the handcrafted features invariant two different poses lightning and viewpoint variation. Learning a similarity metric on the basis of extracted features to determine different images of the same group. 3) and learning these features as well as similarity metrics. Deep learning methods are recent to the developed methods. These methods employed different neural network models like conventional neural network generative adversarial network and recurrent neural network for addressing the problem of person re-identification. Very recently, few researchers have proposed new methods [32][5] on the basis of an unsupervised domain adaptation using a generative adversarial network (GAN). GAN enables the translation of images from the source to the destination domain. They considered the difference between the source and destination domains in addition to the bias between the source and destination camera domains. Thus, it provides motivation for conducting a comprehensive review of deep learning methods specifically GAN based methods for solving person re-identification problem.

This review is organized as follows. In section 2, we present general principles of GAN and its common architecture. Section 3 provides the details person re-identification optimization problem and possible solutions. In Section 4, we review GAN based methods. Section 5 describes different datasets and performance metrics. Finally, Section 6 concludes the review paper and provides directions for future research.

## 2   GAN Preliminaries

Firstly, Goodfellow et al. [12] introduced the adversarial process to learn generative models. The fundamental aspect of GAN is the min-max two-person zero-sum game. In this game, one player takes the advantages at the equivalent loss of the other player. The general architecture of GAN is shown in Figure 1. In general architecture, a generative adversarial network has two types of networks called discriminator and generator denoted as D and G respectively.

1. **The Generator (G):** The G is a network that is used to generate the images using random noise Z. The generated images using noise are recorded as G(z). The input that is commonly a Gaussian noise that is a random point in latent space. Parameters of both the G and D networks are updated iteratively during the training process of GAN. Infected the parameters of HDI remain static while training the G network. The output of G network is labelled as fake distribution and given to D network as an input for determining its class. The error is computed between the output of the discriminator and the label of the sample image. The error is propagated back to update the weights of G network. A few constraints have been imposed on input parameters of G network that can be added in the last layer of this network. In addition, noise can also be added to the hidden layers. There is no limit on the dimensions of input Z of G networks.

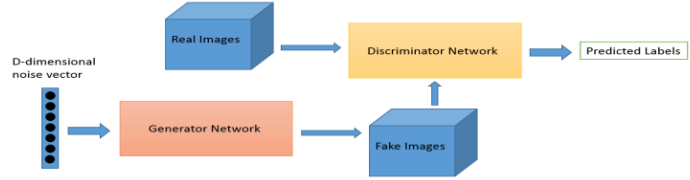

Fig. 1. The general architecture of GAN.

2. **The Discriminator (D):** The D is considered as a discriminant network to determine whether a given image belongs to a real distribution or not. It receives an input image X and produces the output D (x), representing the probability that X belongs to a real distribution. If the output is 1, then it indicates a real image distribution. The output value of D as 0 indicates that it belongs to a fake image distribution. During the training process of the network, G network remains static. D network takes a real image as well as the fake image generated by g network as input to compute the error in its label prediction. The weights of the discriminator D network are updated on the basis of back-propagated error.

The objective function of a two-player minimax game would be as Eq. 1.

$$\mathrm{Min}_G \mathrm{Max}_D V(D, G) = E_{x \sim p_{data}(x)}[\log(d(x))] + E_{z \sim p_g(z)}[\log(1 - D(G(z)))] \quad (1)$$

With the passage of time, several developments have been made to the original architecture of GAN as described below.

### 2.1 Conditional GAN

GANs can be extended to a conditional model if both the G and D networks are conditioned on some extra information y to address the limitation of dependence only on random variables in original model [20]. y could be any kind of auxiliary information, such as class labels or data from other modalities. The conditional information can be added by feeding y into the both the D and G network as an additional input layer. In the G network, the prior input noise $p_z(z)$, and y are combined in joint hidden representation, and the adversarial training framework allows for considerable flexibility in how this hidden representation is composed [20]. In the D network, x and y are presented as inputs and to a D function. The objective function of a two-player minimax game would be as Eq. 2.

$$Min_G Max_D V(D, G) = E_{y,x \sim p_{data}(y,x)}[\log(d(y, x))] + E_{x \sim p_x, z \sim p_z(z)}[\log(1 - D(G(z, x), x))] \quad (2)$$

## 2.2 GAN with auxiliary classifier

Odena et al. [21] proposed a variant of the GAN architecture called as an auxiliary classifier GAN (or AC-GAN). In the ACGAN, every generated sample has a corresponding class label, $c \sim p_c$ in addition to the noise z. G network uses both to generate images $X_{fake} = G(c; z)$. The discriminator gives both a probability distribution over sources and a probability distribution over the class labels, $P(S \mid X); P(C \mid X) = D(X)$. The objective function has two parts: the log likelihood of the correct source, LS, and the log-likelihood of the correct class, $L_C$ as presented in Eqs. 3 and 4.

$$L_s = E[Log P(S = real \mid X_{real})] + E[Log P(S = fake \mid X_{fake})] \quad (3)$$

$$L_s = E[Log P(C = c \mid X_{real})] + E[Log P(C = c \mid X_{fake})] \quad (4)$$

## 2.3 GAN with Encoder

Although GAN can transform a noise vector z into a synthetic data sample G(z), it does not allow inverse transformation. If we treat the noise distribution as a latent feature space for data samples, GAN lacks the ability to map data sample x into latent feature z. In order to allow such mapping, two concurrent works BiGAN [6] and ALI [7] proposed adding an encoder E in the original GAN framework.

Let $\Omega_x$ be the data space and $\Omega_z$ be the latent feature space, the encoder E takes as input and produce a feature vector $E(x) \in \Omega_x$, z as output. The discriminator D is modified to take both a data sample and a feature vector as input to calculate $P(Y \mid x; z)$, where $Y = 1$ indicates the sample is real and $Y = 0$ means the data is generated by G. The objective function here is as per Eq. 5.

$$Min_{G,E} Max_D V(G, E, D) = E_{x \sim p_{data}(x)}[\log(x, E(x))] + E_{z \sim p_g(z)}[\log(1 - D(G(z), z))] \quad (5)$$

## 2.4 GAN with Variational Autoencoder

Larsen et al. [14] found that by jointly training a VAE and a generative adversarial network (GAN) [12]. It can be used the GAN discriminator to measure sample similarity. They achieved this by combining a VAE with a GAN. They proposed to collapse the VAE decoder and the GAN generator into one by letting them share parameters and training them jointly. For the VAE training objective, we replace the typical element-wise reconstruction metric with a feature wise metric expressed in the discriminator. The VAE part regularize the encoder E by imposing a prior of normal distribution (e.g. $z \sim N(0, 1)$), and the VAE loss term is defined in Eq. 6.

$$L_{VAE} = - E_{x \sim q(z|x)}[\log[p(x \mid z)] + D_{KL}(q(z \mid x) k p(x)) \qquad (6)$$

Where $z \sim E(x) = q(z \mid x)$, $x \sim G(z) = p(x \mid z)$ and $D_{KL}$ is the Kullback-Leibler divergence.

## 3 Person Re-identification

Person re-identification is one of the challenging tasks due to various human poses, domain differences, occlusions, etc. [33][26][24][24] [25][2]. It can be performed by using two types of methods, namely (i) similarity measures or learning distance for predicting similarity among two images of a person. [3][36][30][31] and (ii) developing a distinctive signature for representing a person under different camera environment having classification typically on cross-image representation [1] [28] [35]. The researchers in the first type of methods generally utilize several kinds of hand-crafted features like local binary patterns, local maximal occurrence (LOMO), colour histogram, and focus on learning an effective distance/similarity metric for comparing the features. The second type of methods employs deep convolutional neural networks that are very effective in localizing/extracting relevant features to form discriminative representations against view variations.

Very recently, GAN is deriving increasing attention of the researchers for solving person re-identifcation problem. Recently, some researchers used the potential of GAN for aiding person re-identification methods.

## 4 GAN based Person Re-identification Approaches

Several researchers have developed variants of standard GAN for solving person re-identification problem. Different models have been applied to enhance the accuracy of person re-identification using benchmark datasets as described below.

### 4.1 Person Re-identification with Pose Normalization (PN-GAN)

Person re-identification is a very challenging problem because a person's appearance can change drastically across views, due to the changes in various co-variate factors independent of the person's identity. These factors include viewpoint, body configuration, lighting, and occlusion. Among these factors, pose plays an important role in causing a person's appearance changes. Here, pose is defined as a combination of viewpoint and body configuration. It is thus also a cause of self-occlusion. Qian et al. [22] proposed a pose-normalization GAN model (PN-GAN) for alleviating the impact of pose variation. Given a pedestrian image, the model utilized a desirable pose to produce a composite image of the same ID with the initial pose replaced with the desirable pose. Following this, the authors used the pose-normalized images and original images for training the re-identification model to generate two sets of features. In the end, they fused the two types of features for forming final descriptor. As a result, GAN-based data augmentation method enabled the enhancement in generalization ability of re-identification model and solved person re-identification problem from a certain standpoint to a certain extent. In their experiments, the authors used VGG-19 pre-trained on the ImageNet ILSVRC- 2012 dataset to extract the features of each pose images. K-means algorithm was used to cluster the training pose images into canonical poses. The mean pose images of these clusters are then used as the canonical poses. The eight poses obtained on Market-1501 [39]. They used four datasets, Market-1501 [39], CUHK03 [17], DukeMTMC-reID [23] and CUHK01. The results are computed in terms of different ranks of accuracy and mean Average Precision (mAP). Extensive experiments on these four benchmarks showed that their model achieves state-of-the-art performance. This model differs significantly from the previous models [26][37] [38] in that they synthesize realistic whole-body images using the proposed PN-GAN, rather than only focusing on body parts for pose normalization. However, the quality of produced images was comparatively poor leading to fetching noise to the re-identification model.

### 4.2   Person Re-identification with Feature Distilling GAN (FD-GAN)

The issue of pose variation in person images has also been addressed by Ge et al. [11]. The authors proposed a Feature Distilling Generative Adversarial Network (FD-GAN) to learn identity-related and pose-unrelated representations. The proposed system relies on a Siamese structure with multiple novel discriminators on human poses and identities. In addition to the discriminators, they suggested a novel same-pose loss integration that needs the appearance of the same person's produced images to be similar. After learning pose-unrelated person features with pose guidance, no auxiliary pose information and additional computational cost are required during testing. The proposed FD-GAN obtained better performance on three-person re-identification datasets demonstrating its effectiveness and robust feature distilling capability. The authors evaluated their model using Market-1501 [39], CUHK03 [17] and DukeMTMCreID using Top 1 accuracy and mAP metrics and compared with the state-of-the-art person reID methods. The results indicates the superiority of the FD-GAN system. FD-GAN achieved 90.5% top-1 accuracy and 77.7% mAP on the Market-1501 dataset, 92.6% top-1 accuracy and 91.3% mAP on CUHK03 [17] dataset, and 80.0% top-1 accuracy and 64:5% mAP on the DukeMTMCreID dataset, which demonstrated the effectiveness of FD-GAN system.

### 4.3   Cross GAN for cross-view person re-id (cross-GAN)

Dai et al. [4] proposed a cross-modality GANs model to analyze the re-identification between RGB and infrared images. They tackled the challenges of lack of dis-criminative information to re-identify the same person between RGB and infrared modalities, and the difficulty to learn a robust metric for such a large-scale cross-modality retrieval. For handling the lack of insufficient discriminative information issue, the authors developed a cutting-edge generative adversarial training based discriminator to learn discriminative feature representation from different modalities. For handling the issue of large-scale cross-modality metric learning, they integrated both identification loss and cross-modality triplet loss, which minimize inter-class ambiguity while maximizing cross-modality similarity among instances. The proposed model was trained in an end-to-end way using a standard deep neural network framework. They validated their approach using SYSU RGB-IR re-identification benchmark dataset. They reported superior performance in terms of Cumulative Match Characteristic curve (CMC) and Mean Average Precision (mAP) over the method proposed in [33] by at least 12.17% and 11.85% respectively.

### 4.4 Cross Dataset person re-id using GAN (IPGAN)

Liu et al. [19] proposed the Identity IPGAN that ensures the transferred image has a similar style as the style in target camera domain. The method is also able to keep the identity information of images from source domain during the translation. IPGAN consists of a style transfer model $G(x; c)$, a domain discriminator $D_d$ om, and a semantic discriminator $D_s$ em. The construction of IPGAN requires a source training set, the identity labels of source training set, a target training set, and the camera labels of target training set. They trained their model on the translated images by supervised methods and compared their results on Market-1501 [39] and DukeMTMC-reID [23] in terms of Rank 1, 5, 10 of accuracy and mAP metrics. The reported results indicate that the images generated by IPGAN are more suitable for cross-domain person re-identification. They compared the proposed method with the state-of-the-art unsupervised learning methods, hand crafted features like Bag-of-Words(BoW) and local maximal occurrence (LOMO), unsupervised methods like CAMEL, PUL, and UMDL and GAN based methods like PTGAN, SPGAN(+LMP), TJ-AIDL and CamStyle. Their method achieved rank-1 accuracy = 57.2% and the best mAP = 28.0.

### 4.5 Label smoothing regularization method (LSRO)

Zheng et al. [41] proposed the use of a GAN for generating unlabeled pedestrian image samples and employed a CNN sub-model for feature representation learning. As the generated pedestrian images have no labels, through using a label smoothing regularization for outliers (LSRO) method, the model merges the unlabeled GAN images with the real labelled images for training. The major advantage of using LSRO enables the dealing with more training images (outliers) that are located near the real training images in the sample space, and introduce more color, lighting and pose variances to regularize the model. For person re-identification dataset, the image style of different cameras in it may be different that results to the pedestrian images clicked by non-overlapping cameras suffer from intensive changes in background and appearance. The authors used the GAN for generating new pedestrian images with new labels by semi-supervised learning. They denoted the generated images with the LSRO label distribution. On Market-1501 [39], the authors reported rank-1 accuracy of 78.06%, mAP of 56.23%, and reached a rank-1 accuracy of 73.1%, mAP of 77.4% on CUHK03 [17], which are all very competitive. The new idea of the proposed approach was to improve the generalization capability by producing new samples with LSRO label. But, still, new images are too blurry for meeting artificial benchmarks and unable to add into the size of data directly.

### 4.6 Person Transfer GAN to bridge Domain Gap( PTGAN)

The major issues in person re-identification concern the complex scenes and lighting variations, viewpoint and pose changes, and the large number of identities in a camera network. Wei et al. [32] contributed a new dataset called Multi-Scene Multi-Time person ReID dataset (MSMT17) with many important features, e.g., 1) the raw videos are taken by an 15-camera network deployed in both indoor and outdoor scenes, 2) the videos cover a long period of time and present complex lighting variations, and 3) it contains currently the largest number of annotated identities, i.e., 4,101 identities and 126,441 bounding boxes. They authors also revealed that, domain gap commonly exists between datasets, which essentially causes severe performance drop when training and testing on different datasets. This results in that available training data cannot be effectively leveraged for new testing domains. To relieve the expensive costs of annotating new training samples, Wei et al. [32] proposed a system called PTGAN for bridging the domain gap between separate person re-identification datasets. It is inspired by the Cycle-GAN [43]. Different from Cycle-GAN [43], PTGAN considers extra constraints on the person foregrounds to ensure the stability of their identities during transfer. Compared with Cycle-GAN, PTGAN generates high quality person images, where person identities are kept and the styles are effectively transformed. Extensive experimental results on several datasets show PTGAN effectively reduces the domain gap among datasets. The proposed loss for PTGAN is presented in Eq. 7.

$$L_{PTGAN} = L_{Style} + \lambda_1 L_{ID} \qquad (7)$$

Where, $L_{Style}$ denotes the style loss and $L_{ID}$ denotes the identity loss, and $\lambda_1$ is the parameter for the trade-off between two losses. The authors used DukeMTMC-reID [23], Market-1501 [39], CUHK03 [17], PRID datasets fore evaluating PTGAN system in terms of rank 1, 5, 10, and 20 of accuracy and mAP. The results indicates Rank-1 accuracy is constantly improved by 1.9%, 5.1%, and 2.4%, respectively by combining the transferred data from Duke, Market, CUHK03 [17], and MSMT17.

### 4.7 Adversarial open world person re-id (APN)

Li et al. [18] proposed a deep open-set group-based person re-identification method called Adversarial PersonNet (APN) that adopts the adversarial learn-ing strategy to relieve the attack of similar non-target person. This work used target-like samples produced by GAN for attacking the feature extractor and making the extractor can tolerate the attack through discriminative learning. Through the adversarial approach, person re-identification system gets more stabilized while facing the open-world issue. The author evaluated their method using Market-1501 [39], CUHK01, and CUHK03 [17] and proved better performance over the existing methods.

### 4.8 GAN for occluded people in person re-id

Wu et al. [34] also suggested an approach for synthesizing labelled person images automatically and adopting them for increasing the sample number for per identity in datasets. The authors used the block rectangles for occluding the random parts of the persons in the images. They proposed a GAN model for using a paired occlusion and original images to synthesize the de-occluded images that similar but not identical to the original images. Later, they commented on the de-occluded images with the same labels of their corresponding raw image and used them to augment the training samples. They used the augmented datasets to train the baseline model. The experiment results on CUHK03 [17], Market-1501 [39] and DukeMTMC-reID [23] datasets show that the effectiveness of the proposed method in terms of Rank 1, 5 and 10 of accuracy and mAP.

Fabbri et al. [8] proposed a model for handling the issue of occlusion and low resolution of pedestrian attributes using deep generative models (DCGAN). Their model has three sub-networks, for the attribute classification network, the reconstruction network and super-resolution network. For the attribute classification network, the authors used joint global and local parts for final attribute estimation. They utilized ResNet50 to extract the deep features and global-average pooling to obtain the corresponding score. These scores are fused as the final attribute prediction score. For tackling the occlusion and low-resolution problem, they suggested the deep generative adversarial network [12] for generating re-constructed and super-resolution images. Their model used the pre-processed images as input to the multi-label classification network for attribute recognition.

Fulgeri et al. [10] proposed an approach by integrating the existing neural network architectures, namely U-nets and GANs, as well as discriminative attribute classification nets, with an architecture specifically designed to de-occlude people shapes. They trained their network for optimizing a loss function taking into consideration the objectives of generating image a person for a given occluded version as input a) without occlusion b) similar at the pixel level to a completely visible people shape c) capable of conserving similar visual attributes of the original one. The authors evaluated their approach RAP dataset [16] and AiC Dataset [10], compared with state-of-the-art methods and performing the ablation study over each loss employed in terms of five evaluation metrics for the attribute classification task, namely mean Accuracy, Accuracy, Precision, Recall and F1. The authors reported an accuracy=66.23%, Precision =77.85%, Recall=79.71%, F1-measure=78.77% and mAP=78.66% using RAP dataset [16].

The reported results are accuracy=74.87%, Precision =76.80%, Recall=95.43%, F1-measure=85.11% and mAP=91.89% using AiC dataset.

### 4.9 Camera style adaptation for person re-id

Zhong et al. [42] proposed a camera style adaption model for adjusting CNN training. Particularly, the authors used the CycleGAN [43] for transferring the style of images clicked by one camera to another. Given a training image from some camera, the model can generate new images with the style of other cameras. In addition, to alleviating the noise of produced image using CycleGAN, the authors suggested smooth label regularization to the newly generated images. So far, different person re-identification datasets exist domain gap. The gap would lead to serious performance degradation during model training on a given dataset and testing on another dataset. Given that both the real and fake (style-transferred) images have ID labels, the authors used the ID-discriminative embedding (IDE) to train the re-ID CNN model. Using the Softmax loss, IDE regards re-ID training as an image classification task. They used ResNet-50 as backbone and follow the training strategy for fine-tuning on the ImageNet pre-trained model. Different from the IDE proposed in [40], they discard the last 1000-dimensional classi-fication layer and add two fully connected (FC) layers. Theoutput of the first FC layer has 1024 dimensions named as "FC-1024", followed by batch normalization, ReLU and Dropout. The addition "FC-1024" yields improved accuracy. The output of the second FC layer, is C-dimensional, where C is the number of classes in the training set. In their implementation, all input images are resized to 256 ×128.

### 4.10 Image to Image domain adaptation for person re-id (SPGAN)

Deng et al. [5] considered the domain adaptation in person re-identification that task aims at searching for images of the same person to the query. They proposed a heuristic solution, named similarity preserving cycle-consistent generative adversarial network (SPGAN). In their method, SPGAN is only used to improve the first component in the baseline, i.e., image-image translation. They performed image-image translation and re-identification feature learning separately. SPGAN is composed of an Siamese network (SiaNet) and a CycleGAN. Using a contrastive loss, the SiaNet pulls close a translated image and its counterpart in the source, and push away the translated image and any image in the target. They evaluated their methods on two large-scale datasets, i.e., Market-1501 [39] and DukeMTMC-reID [23] in terms of Accuracy ranking and mAP. They proved that SPGAN has better qualify the generated images for domain adaptation and achieve the state-of-the art results on two large-scale person re-ID datasets.

## 5 Benchmark Datasets and Metrics

Several benchmark datasets have been utilized for evaluating person re-identification methods. The significant datasets are described below.

1. Market-1501 [39]: It is collected from 6 different camera views. It has 32,668 bounding boxes of 1,501 identities obtained using a Deformable Part Model (DPM) person detector.
2. CUHK03 [17]: It contains 14,096 images of 1,467 identities, captured by six camera views with 4.8 images for each identity in each camera on average.
3. DukeMTMC-reID [23]: It is constructed from the multi-camera tracking dataset DukeMTMC. It contains 1,812 identities.
4. CUHK01 [15]: It has 971 identities with 2 images per person captured in two disjoint camera views respectively.
5. SYSU RGB-IR Re-ID [27]: It is the first benchmark for cross-modality (RGB-IR) Re-ID, which is captured by 6 cameras, including two IR cameras and four RGB ones. This dataset contains 491 persons with total 287,628 RGB images and 15,792 IR images from four RGB cameras and two IR cameras. The dataset is separated into the training set and the test set, where images of the same person can only appear in either set. And the training set consists of total 32,451 images including 19,659 RGB images and 12,792 IR images. It is a very challenging dataset due to the great differences between two modalities.
6. RAP dataset [16]: It a richly annotated dataset with 41,585 pedestrian samples, each of which is annotated with 72 attributes as well as viewpoints, occlusions and body parts information.
7. AiC Dataset [10]: Attributes in Crowd (AiC) dataset, a novel synthetic dataset for people attribute recognition in presence of strong occlusions. AiC features 125,000 samples, all being a unique person, each of which is automatically labeled with information concerning sex, age etc. The dataset is split into 100,000 samples for training and 25,000 for testing purposes. Each of the 24 attributes is present at least in a 10% of samples which highlight a good balance in terms of labels. The collected samples feature a vast number of different body poses, in several urban scenarios with varying illumination conditions and viewpoints. Skeleton joints are also available for each identity. Joints are additionally labeled with an occlusion flag which tells if the specific body part is directly visible from the camera point of view.

Different researchers using above-cited dataset and performance metrics can be summarized in Table 1.

## 6 Conclusion and Future Directions

Person re-identification based on deep learning approaches is still in the development stage. Presently, these deep learning methods introduced are mostly based on the improvement of their existing basic network models. As per person image characteristics, the establishment of a new deep network model becomes the main research direction. No doubt, these methods have gained better results, it is wasted to adjust the parameters in the training process constantly, and it is easy to occur over-fitting when network structure becomes deep. In addition, improve the computation costs and save time to its maximum extent are the keys to practical. As the number of training samples affects the deep learning models performance, only establishing a large-scale standard dataset can ensure that the trained models still have a good generalization ability under the complex environment. At the same time, as a multi-camera and cross-view recognition application, person re-identification will be applied in reality occasions. However, the existing methods are difficult to ensure the recognition accuracy and the rapidity at the same time. The GAN-based deep model served as a data augmentation technique that can address the limitations of the existing person re-identification problems to a certain extent. But the qualities of most of the synthesized person images produced by the current GAN-based models are not high. These models lead to noise to the original datasets.

Table 1. Summary of Datasets and Performance Metrics.

| Sr | Study | Dataset | Performance Metrics |
|---|---|---|---|
| 1 | PNGAN [22] | Market-1501, CUHK03, DukeMTMC-reID, CUHK01 | Rank 1, mAP |
| 2 | FDGAN [11] | Market-1501, CUHK03, DukeMTMCreID | Top 1 accuracy, mAP |
| 3 | cmGAN [4] | SYSU RGB-IR Re-ID | CMC, mAP |
| 4 | IPGAN [19] | Market-1501, DukeMTMC-reID | Rank 1, 5, 10 of accuracy and mAP |
| 5 | PTGAN [32] | DukeMTMC-reID, Market-1501, CUHK03, PRID | Rank 1, 5, 10, 20 of Accuracy and mAP |
| 6 | DCGAN [8] | RAP | Accuracy, Precision, Recall, F1-measure, mAP |
| 7 | LSRO [41] | Market-1501, CUHK03, DukeMTMC-reID | Rank 1, 5 and 10 of Accuracy, mAP |
| 8 | APN [18] | Market-1501, CUHK01, CUHK03 | True Target Rate (TTR), false target rate (FTR) |
| 9 | Random Occlusion GAN [34] | CUHK03, Market-1501, DukeMTMC-reID | Rank 1, 5 and 10 of accuracy and mAP |
| 10 | U-netsGAN [10] | RAP, AiC | mAP, Accuracy, Precision, Recall, F1-Measure |
| 11 | SPGAN [5] | Market-1501, DukeMTMC-reID | Accuracy ranking and mAP |

Future GAN-based data augmentation must take into consideration for producing more quality samples for the datasets. The existing GAN-based methods have been developed for image-based person re-identification. A promising direction for future research is to using GAN model for generating a sequence of image samples for the video-based identification datasets.